\documentclass{article}

\PassOptionsToPackage{numbers,sort&compress}{natbib}
\usepackage[preprint]{nips_2018}
\usepackage{amsmath, amsfonts, amssymb}
\usepackage{tikz-cd}
\usepackage{units}

\usepackage{microtype}
\usepackage{graphicx}
\usepackage{subfigure}
\usepackage{booktabs} 
\usepackage{makecell}
\usepackage[detect-all]{siunitx}
\usepackage[hyphens]{url}
\usepackage{wrapfig}

\usepackage[utf8]{inputenc} 
\usepackage[T1]{fontenc}    
\usepackage{hyperref}       
\usepackage{url}            
\usepackage{booktabs}       
\usepackage{amsfonts}       
\usepackage{nicefrac}       
\usepackage{microtype}      

\title{Tensor field networks:\\Rotation- and translation-equivariant neural networks for 3D point clouds}

\author{
  Nathaniel Thomas\thanks{Equal contribution.} \\
  Stanford University \\
  Stanford, California, USA \\
  \texttt{ncthomas@stanford.edu} \\
  \And
  Tess Smidt$^{*}$ \\
  University of California, Berkeley \\
  Berkeley, California, USA \\
  Lawrence Berkeley National Lab \\
  Berkeley, California, USA \\
  \texttt{tsmidt@berkeley.edu} \\
  \And
  Steven Kearnes \\
  Google \\
  Mountain View, California, USA \\
  \texttt{kearnes@google.com} \\
  \And
  Lusann Yang \\
  Google \\
  Mountain View, California, USA \\
  \texttt{lusann@google.com} \\
  \And
  Li Li \\
  Google \\
  Mountain View, California, USA \\
  \texttt{leeley@google.com} \\
  \And
  Kai Kohlhoff \\
  Google \\
  Mountain View, California, USA \\
  \texttt{kohlhoff@google.com} \\
  \And
  Patrick Riley \\
  Google \\
  Mountain View, California, USA \\
  \texttt{pfr@google.com} \\
}

\begin{document}

\maketitle

\begin{abstract}
We introduce tensor field neural networks, which are locally equivariant to 3D rotations, translations, and permutations of points at every layer. 3D rotation equivariance removes the need for data augmentation to identify features in arbitrary orientations. Our network uses filters built from spherical harmonics; due to the mathematical consequences of this filter choice, each layer accepts as input (and guarantees as output) scalars, vectors, and higher-order tensors, in the geometric sense of these terms.  We demonstrate the capabilities of tensor field networks with tasks in geometry, physics, and chemistry.
%
%
%
\end{abstract}

\section{Motivation} \label{sec:motivation}

Convolutional neural networks are translation-equivariant, which means that features can be identified anywhere in a given input.  This capability has contributed significantly to their widespread success.

In this paper, we present a family of networks that enjoy richer equivariance: the symmetries of 3D Euclidean space. This includes 3D rotation equivariance (the ability to identify a feature in any 3D rotation and its orientation) and 3D translation equivariance.

Equivariance confers three main benefits:
First, this is more efficient than data augmentation to obtain 3D rotation-invariant output, making computation and training less expensive. This is significantly more important in 3D than in 2D: Without equivariant filters like those in our design, achieving an angular resolution of $\delta$ would require a factor of $\mathcal{O}(\delta^{-1})$ more filters in 2D but $\mathcal{O}(\delta^{-3})$ more filters in 3D.\footnote{This is because the manifold of orthonormal frames at a point in 2D (the group $O(2)$) has dimension 1 and in 3D ($O(3)$) has dimension 3.} Second, a 3D rotation- and translation-equivariant network can identify local features in different orientations and locations with the same filters, which can simplify interpretation. Finally, the network naturally encodes geometric tensors (such as scalars, vectors, and higher-rank geometric objects), mathematical objects that transform predictably under geometric transformations of rotation and translation.  In this paper, the word ``tensor'' refers to {\em geometric} tensors, 
not generic multidimensional arrays.

Our network differs from a traditional convolutional neural network (CNN) in three ways:
\begin{itemize}
\item We operate on point clouds using continuous convolutions. 
Our layers act on 3D coordinates of points and features at those points.
\item We constrain our filters to be the product of a learnable radial function and a spherical harmonic. 
\item Our filter choice requires the structure of our network to be compatible with the algebra of geometric tensors.
\end{itemize}
We call these {\em tensor field networks} because every layer of our network inputs and outputs tensor fields: scalars, vectors, and higher-order tensors at every geometric point in the network. Tensor fields are ubiquitous in geometry, physics, and chemistry, and we expect tensor field neural networks to have applications in each of these areas.

Our initial motivation was to design a universal architecture for deep learning on atomic systems (such as molecules or materials). Tensor field networks could also be used to process 3D images in a rotation- and translation-equivariant way.  We mention other potential applications in Section~\ref{sec:future}.
%
%
%

In this paper, we explain the mathematical conditions that such a 3D rotation- and translation-equivariant network must satisfy, provide several examples of equivariant-compatible network components, and give examples of tasks that this family of networks can perform.

\section{Related work}

Our work builds upon Harmonic Networks~\citep{Worrall_2017_CVPR}, which achieves 2D rotation equivariance using discrete convolutions and filters composed of circular harmonics, and SchNet \citep{MD17Schutt2017}, which presents a rotation-invariant network using continuous convolutions. 
The networks presented in these papers can be emulated by tensor field networks.  However, the mathematics of rotation equivariance in 3D is much more complicated than in 2D because rotations in 3D do not commute; that is, for 3D rotation matrices $A$ and $B$, $AB \ne BA$ in general (see \citet{reisert2009spherical} for more about the mathematics of tensors under 3D rotations). 

Other authors have investigated the problems of rotation equivariance in 2D, such as 
\citet{2017arXiv170101833Z,DBLP:journals/corr/GonzalezVKT16,2017arXiv170508623L}. Most of these work by looking at rotations of a filter; they differ in exactly which rotations and how that orientation information is preserved (or not) as it is passed through the network.

Previous work has dealt with similar issues of invariance or equivariance under particular input transformations.  G-CNNs \citep{group_equivariant_cnns} guarantee equivariance with respect to \emph{finite} symmetry groups (unlike the continuous groups in this work). \citet{s.2018spherical} use spherical harmonics and Wigner $D$-matrices but only address spherical signals (2D data on the surface of a sphere). \citet{2018arXiv180102144K} use tensor algebra to create neural network layers that extend Message Passing Neural Networks~\citep{pmlr-v70-gilmer17a}, but they are permutation group tensors (operating under permutation of the indices of the nodes), not geometric tensors. The networks presented in \citet{qi2016pointnet,qi2017pointnetplusplus} operate on point clouds and use symmetric functions to encode permutation invariance, but these networks do not include rotation equivariance.

Other neural networks have been designed and evaluated on atomic systems using nuclei centered calculations. Many of these use only the pairwise distance between atoms (e.g. SchNet~\citep{2017arXiv170608566S} and the graph convolutional model from~\citet{faber2017prediction}). \citet{ANI1} additionally uses angular information but does not have general equivariance. 

The other major approach to modeling 3D atomic systems is to voxelize the space \citep{DBLP:journals/corr/WallachDH15,NIPS2017_6935,Torng2017}. In general, these are subject to significant expense, no guarantees of smooth transformation under rotation, and edge effects from the voxelization step.

An introduction to the concepts of steerability and equivariance in the context of neural networks can be found in \citet{steerable_cnns}, which focuses on discrete symmetries.  Further discussion of related theory can be found in \citep{kondor2018generalization}, \citep{kondor2018n}, and \citep{cohen2018intertwiners}.

\section{Group representations and equivariance in 3D}\label{sec:equivariance}

A {\em representation} $D$ of a group $G$ is a function from $G$ to square matrices such that for all $g, h\in G$, \[D(g)D(h) = D(gh)\]  A function $\mathcal{L} : \mathcal{X} \rightarrow \mathcal{Y}$ (for vector spaces $\mathcal{X}$ and $\mathcal{Y}$) is {\em equivariant} with respect to a group $G$ and group representations $D^\mathcal{X}$ and $D^\mathcal{Y}$ if for all $g\in G$, 
\[\mathcal{L} \circ D^\mathcal{X}(g) = D^\mathcal{Y}(g) \circ \mathcal{L}\]
{\em Invariance} is a type of equivariance where $D^\mathcal{Y}(g)$ is the identity for all $g$.
We are concerned with the group of symmetry operations that includes isometries of 3D space and permutations of the points.

Composing equivariant networks $\mathcal{L}_1$ and $\mathcal{L}_2$ yields an equivariant network $\mathcal{L}_2 \circ \mathcal{L}_1$ (proof in supplementary material).
Therefore, proving equivariance for each layer of a network is sufficient to prove 
that a whole network is equivariant.

Furthermore, if a network is equivariant with respect to two transformations $g$ and $h$, then it is equivariant to the composition of those transformations $gh$ (by the definition of a representation).  This implies that demonstrating permutation, translation, and rotation equivariance individually is sufficient to prove equivariance of a network to the group (and corresponding representations) containing all combinations of those transformations.  Translation and permutation equivariance will be manifest in our core layer definitions, so we will focus on demonstrating rotation equivariance.

Tensor field networks act on points with associated features.  A layer $\mathcal{L}$ takes a finite set $S$ of vectors in $\mathbb{R}^3$ and a vector in $\mathcal{X}$ at each point in $S$ and outputs a vector in $\mathcal{Y}$ at each point in $S$, where $\mathcal{X}$ and $\mathcal{Y}$ are vector spaces.  We write this as
\[\mathcal{L}(\vec{r}_a, x_a) = (\vec{r}_{a}, y_{a})\]
where $\vec{r}_a \in \mathbb{R}^3$ are the point coordinates and $x_a\in \mathcal{X}$, $y_{a}\in \mathcal{Y}$ are the feature vectors ($a$ being an indexing scheme over the points in $S$).  
This combination of $\mathbb{R}^3$ with another vector space can be written as $\mathbb{R}^3 \oplus \mathcal{X}$, where $\oplus$ refers to concatenation. 
%
%
%

We now describe the conditions on $\mathcal{L}$ for equivariance with respect to different input transformations.

\subsection{Permutation equivariance}
\[\text{{\em Condition:}} \quad \mathcal{L} \circ \mathcal{P}_\sigma = \mathcal{P}_\sigma \circ \mathcal{L}\]
where $\mathcal{P}_\sigma(\vec{r}_a, x_a) := (\vec{r}_{\sigma(a)}, x_{\sigma(a)})$ and $\sigma$ permutes the points to which the indices refer.

All of the layers that we will introduce in Section~\ref{sec:layers} are manifestly permutation-equivariant because we only treat point clouds as a {\em set} of points, never requiring an imposed order like in a list.  In our implementation, points have an array index associated with them, but this index is only ever used in a symmetric way.
\clearpage
\subsection{Translation equivariance} 
\[\text{{\em Condition:}} \quad \mathcal{L} \circ \mathcal{T}_{\vec{t}} = \mathcal{T}_{\vec{t}} \circ \mathcal{L}\]
where $\mathcal{T}_{\vec{t}}(\vec{r}_a, x_{a}) := (\vec{r}_a + \vec{t}, x_{a})$.  This condition is analogous to the translation equivariance condition for CNNs.

All of the layers in Section~\ref{sec:layers} are manifestly translation-equivariant because we only ever use differences between two points $\vec{r}_i - \vec{r}_j$ (for indices $i$ and $j$).

\subsection{Rotation equivariance}

The group of (proper) 3D rotations is called $SO(3)$, a manifold that can be parametrized by 3 numbers (see~\citet{goodman1998representations}).  Let $D^\mathcal{X}$ be a representation of $SO(3)$ on a vector space $\mathcal{X}$ (and $D^{\mathcal{Y}}$ on $\mathcal{Y}$).  Acting with $g \in SO(3)$ on $\vec{r} \in \mathbb{R}^3$ we write as $\mathcal{R}(g)\vec{r}$, and acting on $x \in \mathcal{X}$ gives~$D^{\mathcal{X}}(g)x$.
\begin{equation}\label{eq:rot_equiv}
\begin{split}
&\text{{\em Condition:}} \\ 
&\mathcal{L} \circ \left[\mathcal{R}(g) \oplus D^{\mathcal{X}}(g)\right] = \left[\mathcal{R}(g) \oplus D^{\mathcal{Y}}(g)\right] \circ \mathcal{L}
\end{split}
\end{equation}
where $\left[\mathcal{R}(g) \oplus D^{\mathcal{X}}(g)\right]\left(\vec{r}_a, x_a\right) = \left(\mathcal{R}(g)\vec{r}_a, D^\mathcal{X}(g) x_a \right)$.
(For layers in this paper, only the action of $D^\mathcal{Y}(g)$ on $\mathcal{Y}$ that will be nontrivial, so we will use a convention of omitting the $\mathbb{R}^3$ layer output in our equations.)

We attain local rotation equivariance by restricting our convolution filters to a particular form.  The features have different types corresponding to whether they transform as scalars, vectors, or higher tensors.

We decompose representations into irreducible representations to simplify our analysis.  The irreducible representations of $SO(3)$ have dimensions $2l + 1$ for $l \in \mathbb{N}$ (including $l=0$) and are unitary.  We will use the term ``rotation order'' to refer to $l$ in this expression.  The rotation orders $l=0, 1, 2$ correspond to scalars, vectors in 3-space, and symmetric traceless matrices, respectively.  

The group elements are represented by $D^{(l)}$, which are called {\em Wigner $D$-matrices} (see~\citet{gilmore2008lie}); they map elements of $SO(3)$ to $(2l+1)\times(2l+1)$-dimensional matrices.  For scalars and 3-space vectors, the (real) Wigner $D$-matrices are 
\[D^{(0)}(g) = 1 \qquad \text{and} \qquad D^{(1)}(g) = \mathcal{R}(g).\] 

\section{Tensor field network layers}\label{sec:layers}

The input and output of each layer of a tensor field network is a finite set $S$ of points in $\mathbb{R}^3$ and a vector in a representation of $SO(3)$ 
associated with each point.  

\begin{wrapfigure}{R}{0.5\columnwidth}
\centering
\includegraphics[width=0.5\columnwidth]{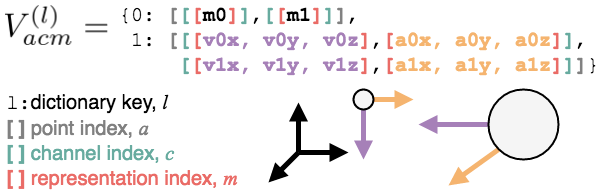}
\caption{Example of $V_{acm}^{(l)}$ representing two point masses with velocities and accelerations. Colored brackets indicate the $a$ (point), $c$ (channel), and $m$ (representation) indices. $(l)$ is the key for the $V_{acm}^{(l)}$ dictionary of feature tensors.}
\label{fig:vacm}
\vspace{-0.2in}
\end{wrapfigure}

We decompose this representation into irreducible representations.  In general, there are multiple instances of each $l$-rotation-order irreducible representations.  These are analogous to what is typically called the ``depth'' of a convolution in a standard CNN, so we will refer to these different instances as {\em channels}.  We implement this object $V^{(l)}_{acm}$ as a dictionary with key $l$ (the rotation order) of multidimensional arrays each with shapes $[|S|, n_l, 2l + 1]$ (where $n_l$ is the number of channels) corresponding to $\texttt{[point index, channel index, representation index]}$. See Figure~\ref{fig:vacm} for an example of how to encode a simple system in this notation.

We will define three tensor field network layers and prove that they are equivariant.  All of these layers will be manifestly permutation-invariant and translation-equivariant, so to prove that a layer is equivariant, we only have to prove rotation equivariance for an arbitrary rotation order.  This requires showing that when the point cloud rotates and the input features are transformed as tensors, the output features also transform as tensors.

\subsection{Point convolution}

This layer is the core of our network and generalizes the point convolutions in \citep{MD17Schutt2017}.
At each layer, we have an input structure $V^{(l)}_{acm}$. The point convolution performs the same operation for each point $a$, taking all the other points (both their relative locations and values of $V$) as input.
In the following, $\vec{r}$ denotes the 3-space vector of an input point to the convolution relative to the convolution center, $\hat{r}$ is $\vec{r}$ normalized to unit length, and $r$ is the length of $\vec{r}$.

Note that our design is strictly more general than standard convolutional neural networks, which can be treated as a point cloud 
of a grid of points with regular spacing.

\subsubsection{Spherical harmonics and filters}

The spherical harmonics $Y^{(l)}_m$ are functions from the points on a sphere to the complex or real numbers (where $l$ is a non-negative integer and $m$ is be any integer between $-l$ and $l$ inclusive). 
The (real) spherical harmonics for $l=0$ (scalars) and $l=1$ (3-space vectors) are 
\[Y^{(0)}(\hat{r}) \propto 1 \qquad \text{and} \qquad Y^{(1)}(\hat{r}) \propto \hat{r}.\]

These functions are equivariant to $SO(3)$; that is, for all $g \in SO(3)$ and $\hat{r}$,
\[Y_m^{(l)}(\mathcal{R}(g)\hat{r}) = \sum_{m'}D_{mm'}^{(l)}(g)Y_{m'}^{(l)}(\hat{r}).\]

To design a rotation-equivariant point convolution, we want rotation-equivariant filters. For our filters to be rotation-equivariant, we restrict them to the following form:
\begin{equation}\label{eq:filter}
F_{cm}^{(l_f, l_i)}(\vec{r}) = R^{(l_f, l_i)}_c(r)Y_m^{(l_f)}(\hat{r})
\end{equation}
(where $l_i$ and $l_f$ are non-negative integers corresponding to the rotation order of the input and the filter, respectively and $R^{(l_f, l_i)}_c : \mathbb{R}_{\ge 0} \rightarrow \mathbb{R}$ are learned functions, which contain most of the parameters within a tensor field network).  Filters of this form inherit the transformation property of spherical harmonics under rotations because $R(r)$ is a scalar in $m$.  This choice of filter restriction is analogous to the use of circular harmonics in \citet{Worrall_2017_CVPR} (though we do not have an analog to the phase offset because of the non-commutativity of $SO(3)$).

\subsubsection{Combining representations using tensor products}

Our filters and layer input each inhabit representations of $SO(3)$ (that is, they both carry $l$ and $m$ indices). In order to produce output that we can feed into downstream layers, we need to combine the layer input and filters in such a way that the output also transforms appropriately (by inhabiting a representation of $SO(3)$).

A {\em tensor product of representations} is a prescription for combining two representations $D^\mathcal{X}$ and $D^\mathcal{Y}$ to get another representation $D^\mathcal{X} \otimes D^\mathcal{Y}$ over the vector space $\mathcal{X} \otimes \mathcal{Y}$.  The crucial property of the tensor product is that it is equivariant:
\[D^{\mathcal{X}} \otimes D^{\mathcal{Y}} = D^{\mathcal{X}\otimes \mathcal{Y}}\]

Now consider the tensor product of two representations of orders $l_1$ and $l_2$ (where we use the usual notational convention of using just $l$ to refer to the $(2l + 1)$-dimensional vector space that represents rotation order $l$). For the irreducible representations with $u^{(l_1)} \in l_1$ and $v^{(l_2)} \in l_2$, $u^{(l_1)} \otimes v^{(l_2)} \in l_1 \otimes l_2$ can be calculated using {\em Clebsch-Gordan coefficients} (see~\citet{griffiths2016introduction}) (denoted with $C$): 
\[(u \otimes v)^{(l)}_{m} = \sum_{m_1=-l_1}^{l_1} \sum_{m_2=-l_2}^{l_2} C^{(l, m)}_{(l_1, m_1) (l_2, m_2)} u^{(l_1)}_{m_1} v^{(l_2)}_{m_2}\]
This tensor product produces non-zero values only for $l$ between $|l_1 - l_2|$ and $(l_1 + l_2)$ inclusive ($m$ is any integer between $-l$ and $l$ inclusive).  The (real) Clebsch-Gordan coefficients for $l=0$ and $l=1$ are just the familiar ways to combine scalars and vectors: 
For $1\otimes 1 \rightarrow 0$ and $1\otimes 1 \rightarrow 1$,
\[C^{(0,0)}_{(1,i)(1,j)} \propto \delta_{ij} \qquad \qquad
C^{(1,i)}_{(1,j)(1,k)} \propto \epsilon_{ijk}\] which are  the dot and cross products for 3D vectors, respectively.  The $0 \otimes 0 \rightarrow 0$ case is just regular multiplication of two scalars, and $0 \otimes 1 \rightarrow 1$ and $1 \otimes 0 \rightarrow 1$ corresponds to scalar multiplication of a vector.
%
%
%
%
%

\subsubsection{Layer definition}

A given input inhabits one representation, a filter inhabits another, and together these produce outputs at possibly many rotation orders.
We can put everything together into our pointwise convolution layer definition:

\begin{equation} \label{eq:conv_def}
\mathcal{L}^{(l_o)}_{acm_o}\big(\vec{r}_a, V^{(l_i)}_{acm_i}\big) := \sum_{m_f, m_i} C_{(l_f, m_f) (l_i, m_i)}^{(l_o, m_o)} \sum_{b\in S} F^{(l_f, l_i)}_{c m_f}(\vec{r}_{ab})V^{(l_i)}_{bcm_i}
\end{equation}
(where $\vec{r}_{ab} := \vec{r}_a - \vec{r}_b$ and the subscripts $i$, $f$, and $o$ denote the representations of the input, filter, and output, respectively).  A point convolution of an $l_f$ filter on an $l_i$ input yields outputs at $2\min(l_i, l_f) + 1$ different rotation orders $l_o$ (one for each integer between $|l_i - l_f|$ and $(l_i + l_f)$, inclusive), though in designing a particular network, we may choose not to calculate or use some of those outputs.

The equivariance of the filter $F$ (equation~\ref{eq:filter})and the equivariance of the Clebsch-Gordan coefficients together imply that point convolutions are equivariant (detailed proof in supplementary material).  This aspect of the design 
is the core of our result in this paper.



\subsection{Self-interaction}\label{sec:self}

We follow \citet{2017arXiv170608566S} in using point convolutions to scale feature vectors elementwise and using self-interaction layers to mix together the components of the feature vectors at each point.  Self-interaction layers are analogous to 1x1 convolutions, and they act like $l=0$ (scalar) filters:
\[\sum_{c'} W^{(l)}_{cc'}V^{(l)}_{ac'm}\]
In general, each rotation order has different weights because there may be different numbers of channels corresponding to that rotation order.  However, the same weights are used for every $m$ for a given order; this is essential to maintain equivariance. For $l=0$, we may also use biases. 

The $D$-matrices commute with the weight matrix $W$ because $W$ has no representation index $m$, so this layer is equivariant for $l > 0$.  Equivariance for $l=0$ is straightforward because $D^{(0)} = 1$.

\subsection{Nonlinearity}

Our nonlinearity layer acts as a scalar transform in the $l$ spaces (that is, along the $m$ dimension).  
For $l=0$ channels, we can use \[\eta^{(0)}\big(V^{(0)}_{ac} + b^{(0)}_c\big) \qquad \text{and}\qquad \eta^{(l)}\big(\|V\|^{(l)}_{ac} + b^{(l)}_c \big)V^{(l)}_{acm} \qquad \text{where} \qquad \|V\|^{(l)}_{ac} := \sqrt{\sum_m |V^{(l)}_{acm}|^2}\] for some functions $\eta^{(l)} : \mathbb{R} \rightarrow \mathbb{R}$ (which can be different for each $l$) and biases $b^{(l)}_c$. 
Note that \[\|D(g)V\| = \|V\|\] because $D$ is a unitary representation.  Therefore, this layer is a scalar transform in the representation index $m$, so it is equivariant.



\section{Demonstrations}
  
We chose demonstrations that both convey the power of 3D equivariance and the flexibility of our framework and are simple enough to serve as clear and unambiguous demonstrations.  Each of these tasks is either unnatural or impossible in existing models.

The largest task that we describe here uses dozens of points, but tensor field networks can scale to over a thousand points on a standard GPU.  We have successfully trained tensor field networks on large standard datasets, including QM9 energy~\cite{QM9ramakrishnan2014quantum}, MD17 molecular dynamics forces~\cite{Chmielae1603015}, and ModelNet40~\cite{DBLP:conf/cvpr/WuSKYZTX15} classification.
 

In these tasks, we used radial functions and nonlinearities identical to those used in \citet{MD17Schutt2017}.  We used radial basis functions composed of Gaussians, and two fully connected layers are applied to this basis vector. 
We implemented our models in TensorFlow \citep{tensorflow2015-whitepaper}, and our code is available at \url{https://github.com/tensorfieldnetworks/tensorfieldnetworks}.

\subsection{Geometry: shape classification}

{\em Network type:} $0 \rightarrow 0$

\begin{wrapfigure}{R}{0.5\textwidth}
\centering
\includegraphics[width=0.5\textwidth]{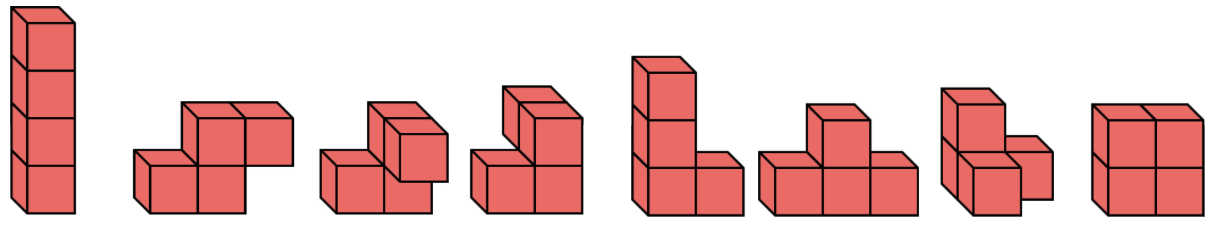}
\caption{3D Tetris shapes. Blocks correspond to single points. The third and fourth shapes from the left are mirrored versions of each other.}
\label{fig:tetris_shapes}
\end{wrapfigure}

Rotation equivariance eliminates the need for rotational data augmentation.  We demonstrate this with the following task:  During training, we input to the network a dataset of shapes in a single orientation, and it learns to classify which shape it has seen.  We then test the network with shapes from the same dataset that have been rotated and translated randomly.  Our network automatically performs as well on this test dataset as it does on the training dataset.

A toy dataset is sufficient to demonstrate this capability; we use 8 shapes that we call {\em 3D Tetris}, shown in Figure \ref{fig:tetris_shapes}, which our network learns to classify with perfect accuracy.  We have also trained a tensor field network on ModelNet40 as in \citep{qi2016pointnet} but without needing rotational data augmentation.  We achieved decent (though not state-of-the-art) accuracy with tensor field networks, but, in contrast, without data augmentation during training, PointNet's performance is roughly random classification accuracy on a test set that has been randomly rotated and translated.

\begin{figure}[t]
\centering
\includegraphics[width=\textwidth]{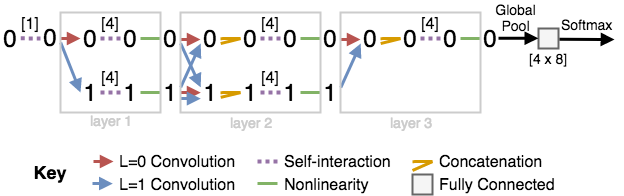}
\caption{Network diagrams for shape classification task showing how information flows between tensors of different order. Clebsch-Gordan tensors are implied in the arrows indicating convolutions. The numbers above the self-interactions indicate the number of channels. Individual convolutions, indicated by arrows, each produce a separate tensor, and concatenation is performed after the convolutions.}
\label{fig:net_diagrams}
\end{figure}

We use a 3-module network that includes the following for every module: all possible paths with $l=0$ and $l=1$ convolutions, concatenation, a self-interaction layer, and a rotation-equivariant nonlinearity. We only use the $l=0$ output of the network since the shape classes are invariant under rotation and hence scalars. 
To get a classification from the $l=0$ output of the network, we sum over the feature vectors of all points. This global pooling operation is equivariant because $D$-matrices act linearly.  This network is depicted in Figure \ref{fig:net_diagrams}.

There are two shapes in 3D Tetris that are mirrors of each other.  Any network that relies solely upon distances (such as SchNet \citep{MD17Schutt2017}) or angles between points (such as ANI-1 \citep{ANI1}) cannot distinguish these shapes, but ours can.


\subsection{Physics: vectors and tensors in classical mechanics}

{\em Network types:} $0 \rightarrow 1$ and $0 \rightarrow 0 \oplus 2$

To demonstrate the usefulness of different tensor field network output types, we train networks to calculate acceleration vectors of point masses under Newtonian gravity and the moment of inertia tensor at a specific point of a collection of point masses. These tasks only require a single layer of point convolutions to demonstrate (though they can also be learned by more general networks), and we can check the learned radial functions against the analytical solutions.

The acceleration of a point mass in the presence of other point masses according to Newtonian gravity is given by
\[\vec{a}_p = -G_N\sum_{n\neq p} \frac{m_n}{r_{np}^2} \hat{r}_{np}\] 
where we define $\vec{r}_{np} := \vec{r}_n - \vec{r}_p$.  (We choose units where $G_N = 1$, for simplicity.)

The moment of inertia tensor is used in classical mechanics to calculate angular momentum.  Objects generally have different moments of inertia depending on which axis they rotate about, and this is captured by the moment of inertia tensor (see~\citet{landau1976mechanics}):
\[I_{ij} = \sum_p m_p \Big[ (\vec{r}_p \cdot \vec{r}_p)\delta_{ij} - (\vec{r}_p)_i (\vec{r}_p)_j \Big]\]

The moment of inertia tensor is a symmetric tensor, so it can be encoded using a $0 \oplus 2$ representation. 

For both of these tasks, we input to the network a set of random points with associated random masses. 
For the moment of inertia task, we also designate a different special point at which we want to calculate the moment of inertia tensor. 

For learning Newtonian gravity, we use a single $l=1$ convolution with 1 channel; for learning the moment of inertia tensor, we use a single layer comprised of $l=0$ and $l=2$ convolutions with 1 channel each. 
We get excellent agreement with the Newtonian gravity inverse square law 
and 
the moment of inertia tensor radial functions. 
Further information about this demonstration is included in the supplementary material.  
\subsection{Chemistry: toward geometrically generating molecular structures}

{\em Network type:} $0 \rightarrow 0 \oplus 1$

In this task, we randomly remove a point from a point cloud and ask the network to replace that point.  This is a first step toward general isometry-equivariant generative models for 3D point clouds.  We train on molecular structures 
because 
precise positions are important for chemical behavior.

\begin{wrapfigure}{L}{0.5\columnwidth}
\centering
\includegraphics[width=0.5\columnwidth]{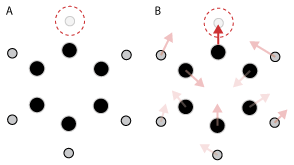}
\caption{A hypothetical example input and output of the missing point network. (A) A benzene molecule with hydrogen removed (B) The relative output vectors produced by the network, with arrows shaded by the associated probabilities.}
\label{fig:benzene_votes}
\vspace{-0.1in}
\end{wrapfigure}

We output an array of scalars, a special scalar, and one vector at each point.  The vector, $\vec{\delta}_a$, indicates where the missing point should be relative to the starting point $\vec{r}_a$ (both direction and distance) and the array of scalars indicates the point type.  The special scalar is used as a probability $p_a$ measuring the confidence in that point's vote.  We aggregate votes for location using a weighted sum:
\[\sum_a p_a (\vec{r}_a + \vec{\delta}_a)\]
(See the supplementary material for a proof that this is translation-equivariant.)  This scheme is illustrated in Figure~\ref{fig:benzene_votes}.

We trained on structures in the QM9 dataset, a collection of~134,000 molecules with up to nine heavy atoms (the elements C, N, O, F).


In a single epoch, 
we train on each molecule in the training set 
with one randomly selected atom deleted. 
We then test the network on a random selection of 
other molecules, 
making a prediction for every possible atom that could be removed. 

\begin{wraptable}{R}{8cm}
\vspace{-0.1in}
\caption{Performance on missing point task}
\vspace{10pt}
\label{table:missing_point}
\centering
\begin{tabular}{ l c c c }
\toprule
Atoms  & {\makecell{Number of \\ predictions}} & {\makecell{Accuracy (\%) \\ ($\le 0.5$~\AA \\ and atom type)}} & {\makecell{Distance \\ MAE in \AA}} \\
\midrule
5-18 & 15863 & 91.3  (train) & 0.16 \\
19  &  19000 & 93.9  (test) & 0.14\\
23  &  23000 &  96.5  (test) & 0.13\\
25-29  & 25356 & 97.3  (test) & 0.16\\
\bottomrule
\end{tabular}
\vspace{-0.35in}
\end{wraptable}

After training for 225 epochs on a dataset containing only 1000 molecules (each with 5-18 atoms), we see excellent generalization to the test sets, which include larger molecules. Further details can be found in the supplementary material.


\section{Future work}\label{sec:future}

We have explained the theory of tensor field networks and demonstrated some of their capabilities.

We expect that tensor field networks will be the right model for learning a wide range of phenomena:  In the context of atomic systems, we intend to train networks to predict properties of large and heterogeneous systems, learn molecular dynamics, calculate electron densities (as inputs to density functional theory algorithms), and hypothesize new stable structures.  Ultimately, we hope to design new useful materials, drugs, and chemicals.

For more general physics, we see potential applications in modeling complex fluid flows, analyzing detector events in particle physics experiments, and studying configurations of stars and galaxies.  We see other applications in 3D perception, robotics, computational geometry, and bioimaging.



\bibliography{main}

\clearpage

\renewcommand\thesection{\Alph{section}}
\setcounter{section}{0}
\renewcommand{\thefigure}{S\arabic{figure}}
\renewcommand{\thetable}{S\arabic{table}}
\setcounter{figure}{0}
\setcounter{table}{0}

\section{Proofs of general equivariance propositions}
\label{sec:proof_equivariance}

For equivariant $\mathcal{L}$, the following diagram is commutative for all $g\in G$:
\[\begin{tikzcd}
\mathcal{X} \arrow[r, "\mathcal{L}"] \arrow[d, "D^\mathcal{X}(g)"] & \mathcal{Y} \arrow[d, "D^\mathcal{Y}(g)"] \\
\mathcal{X} \arrow[r, "\mathcal{L}"]& \mathcal{Y}
\end{tikzcd}\]

If a function is equivariant with respect to two transformations $g$ and $h$, then it is equivariant to the composition of those transformations: 
\[
\begin{split}\mathcal{L}(D^\mathcal{X}(gh)x) &= \mathcal{L}\big(D^\mathcal{X}(g) D^\mathcal{X}(h) x\big) \\
&= D^\mathcal{Y}(g)\mathcal{L}\big(D^\mathcal{X}(h) x\big) \\
&= D^\mathcal{Y}(gh) \mathcal{L}(x)
\end{split}
\] 
for all $g, h \in G$ and $x\in \mathcal{X}$; that is, the following diagram is commutative:
\[\begin{tikzcd}
\mathcal{X} \arrow[r, "\mathcal{L}"] \arrow[d, "D^\mathcal{X}(g)"] & \mathcal{Y} \arrow[d, "D^\mathcal{Y}(g)"] \\
\mathcal{X} \arrow[r, "\mathcal{L}"] \arrow[d, "D^\mathcal{X}(h)"] & \mathcal{Y} \arrow[d, "D^\mathcal{Y}(h)"] \\
\mathcal{X} \arrow[r, "\mathcal{L}"] & \mathcal{Y}
\end{tikzcd}\]

Composing equivariant functions $\mathcal{L}_1 : \mathcal{X} \rightarrow \mathcal{Y}$ and $\mathcal{L}_2 : \mathcal{Y} \rightarrow \mathcal{Z}$ yields an equivariant function $\mathcal{L}_2 \circ \mathcal{L}_1$: 
\[
\begin{split}
\mathcal{L}_2(\mathcal{L}_1(D^\mathcal{X}(g) x)) &= \mathcal{L}_2(D^\mathcal{Y}(g)\mathcal{L}_1(x)) \\
&= D^\mathcal{Z}(g) \mathcal{L}_2(\mathcal{L}_1(x))
\end{split}
\]
That is, the following is commutative:
\[\begin{tikzcd}
\mathcal{X} \arrow[r, "\mathcal{L}_1"] \arrow[d, "D^\mathcal{X}(g)"] & \mathcal{Y} \arrow[r, "\mathcal{L}_2"] \arrow[d, "D^\mathcal{Y}(g)"] 
& \mathcal{Z} \arrow[d, "D^\mathcal{Z}(g)"] \\
\mathcal{X} \arrow[r, "\mathcal{L}_1"] & \mathcal{Y} \arrow[r, "\mathcal{L}_2"] & \mathcal{Z}
\end{tikzcd}\]

\section{Motivating point convolutions}

We can represent input as a continuous function that is non-zero at a finite set of points (using Dirac $\delta$ functions): \[V(\vec{t}) = \sum_{a \in S} V_a \delta(\vec{t} - \vec{r}_a)\]

A point convolution is then equivalent to applying an integral transform with kernel \[F(\vec{t} - \vec{s}) \sum_{a \in S} \delta(\vec{t} - \vec{r}_a)\] for some function $F$.
This transform yields
\begin{align*}\mathcal{L}(\vec{t}) &= \int d^3\vec{s} F(\vec{t} - \vec{s}) \sum_{a \in S} \delta(\vec{t} - \vec{r}_a) \sum_{b \in S} V_b \delta(\vec{s} - \vec{r}_b) \\
&= \sum_{a \in S} \delta(\vec{t} - \vec{r}_a)\sum_{b\in S} F(\vec{r}_a - \vec{r}_b) V_b \\
&= \sum_{a\in S} \delta(\vec{t} - \vec{r}_a)\mathcal{L}_a
\end{align*}
where we define
\[\mathcal{L}_a := \sum_{b\in S} F(\vec{r}_a - \vec{r}_b) V_b\]
as in the main text.

\section{Proof of equivariance of point convolution layer}
\label{sec:proof_point_conv}

\begin{figure}[ht]
\vskip 0.2in
\begin{center}
\centerline{\includegraphics[width=\columnwidth]{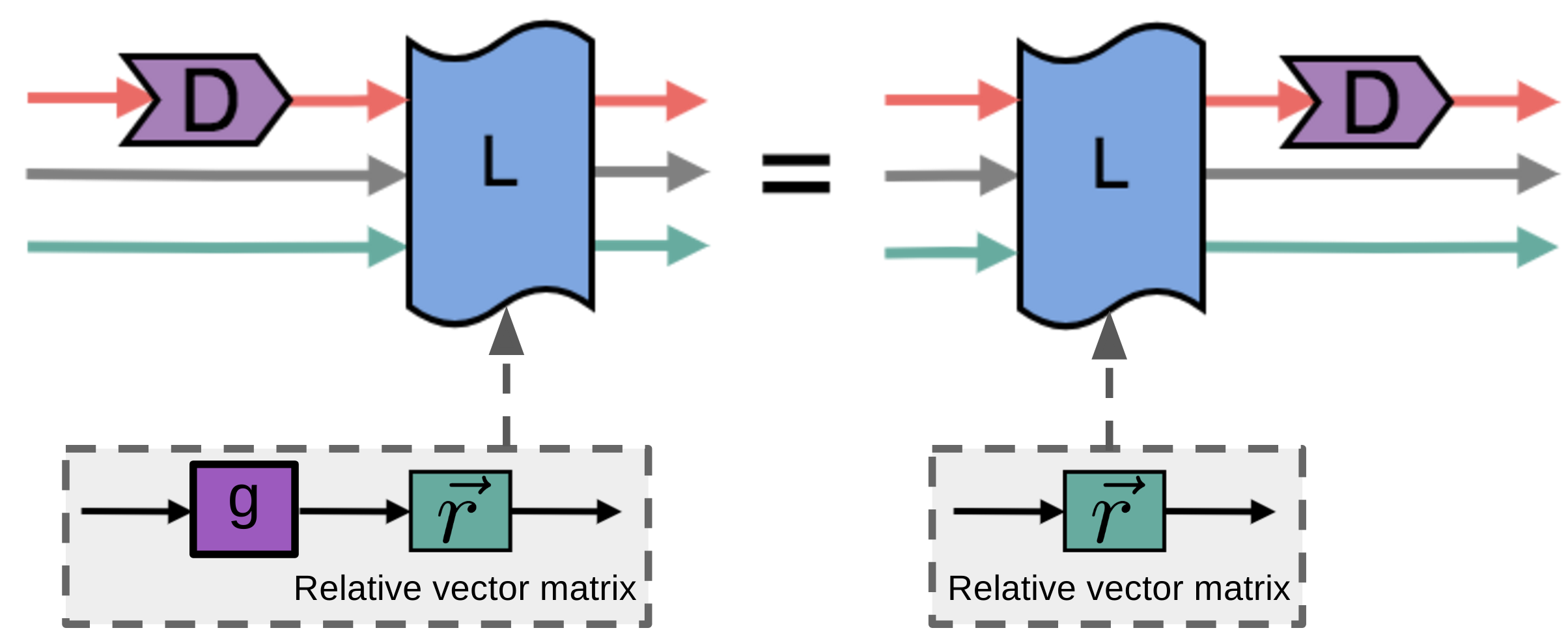}}
\caption{Condition for layer rotation equivariance}
\label{fig:layer_equiv}
\end{center}
\vskip -0.2in
\end{figure}

Under a rotation $\vec{r}_a \mapsto \mathcal{R}(g)\vec{r}_a$, we know that $\vec{r}_{ab} \mapsto \mathcal{R}(g)\vec{r}_{ab}$ and 
\begin{equation}\label{eq:filter_equiv}
F^{(l_f, l_i)}_{cm}(\mathcal{R}(g)\vec{r}_{ab}) = \sum_{m'} D^{(l_f)}_{mm'}(g) F^{(l_f, l_i)}_{cm'}(\vec{r}_{ab})
\end{equation}
because of the transformation properties of the spherical harmonics $Y^{(l)}_m$.

\begin{figure}[ht]
\vskip 0.2in
\begin{center}
\centerline{\includegraphics[width=\columnwidth]{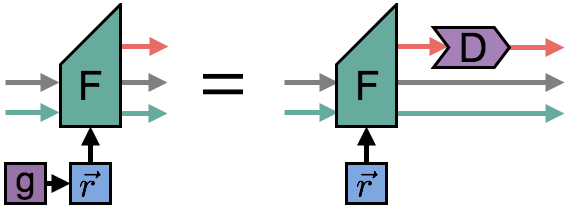}}
\caption{Filter equivariance equation}
\label{fig:filter_equiv}
\end{center}
\vskip -0.2in
\end{figure}

\begin{figure}[ht]
\vskip 0.2in
\begin{center}
\centerline{\includegraphics[width=\columnwidth]{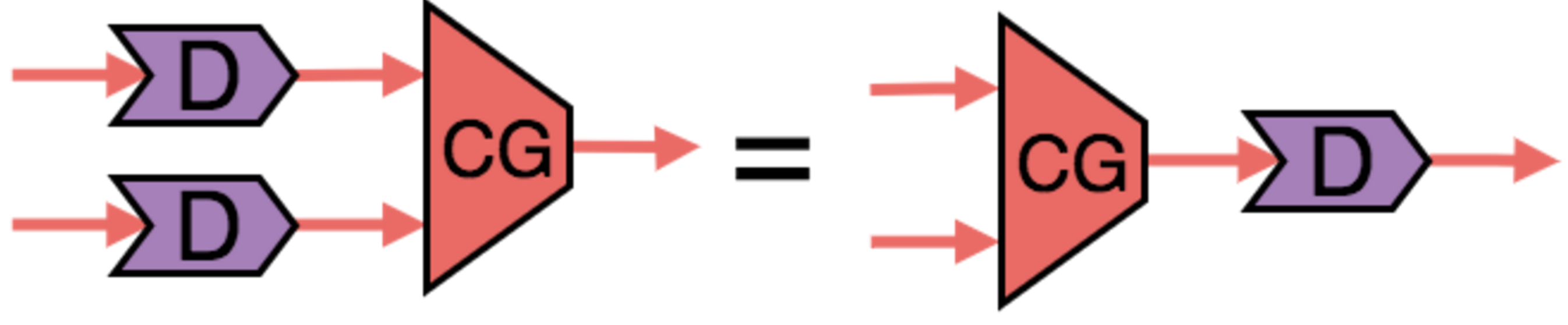}}
\caption{Equivariance of Clebsch-Gordan coefficients.  Note that each $D$ may refer to a different irreducible representation.}
\label{fig:cg}
\end{center}
\vskip -0.2in
\end{figure}

The crucial property of the Clebsch-Gordan coefficients that we need to prove equivariance of this layer is
\begin{align} \label{eq:CDD_DC}
\begin{split}
\sum_{m_1', m_2'} &C_{(l_1, m_1') (l_2, m_2')}^{(l_0, m_0)} D^{(l_1)}_{m_1' m_1}(g) D^{(l_2)}_{m_2' m_2}(g)  \\
&= \sum_{m_0'} D^{(l_0)}_{m_0 m_0'}(g) C_{(l_1, m_1) (l_2, m_2)}^{(l_0, m_0')}
\end{split}
\end{align}
(see Fig \ref{fig:cg} and, for example, \citet{reisert2009spherical}).

Then
\begin{align*}
&\mathcal{L}^{(l_O)}_{acm_O}\Big(\mathcal{R}(g)\vec{r}_a, \sum_{m_I'} D^{(l_I)}_{m_I m_I'}(g) V^{(l_I)}_{acm_I'}\Big) \\
&= \sum_{m_F, m_I} C_{(l_F, m_F) (l_I, m_I)}^{(l_O, m_O)} \sum_{b\in S} F^{(l_F, l_I)}_{c m_F}(\mathcal{R}(g)\vec{r}_{ab}) \sum_{m_I'} D^{(l_I)}_{m_I m_I'}(g) V^{(l_I)}_{bcm_I'} \\
&=\sum_{m_F, m_I} C_{(l_F, m_F) (l_I, m_I)}^{(l_O, m_O)} \sum_{b\in S} 
\left(\sum_{m_F'}D^{(l_F)}_{m_F m_F'}(g)F^{(l_F, l_I)}_{cm_F'}(\vec{r}_{ab})\right) \left(\sum_{m_I'}D^{(l_I)}_{m_I m_I'}(g)V^{(l_I)}_{bcm_I'}\right) \\
&= \sum_{m_O'} D^{(l_O)}_{m_O m_O'}(g) \sum_{m_F, m_I} C_{(l_F, m_F) (l_I, m_I)}^{(l_O, m_O')} \sum_{b\in S} F^{(l_F, l_I)}_{c m_F}(\vec{r}_{ab})V^{(l_I)}_{bcm_I} \\
&= \sum_{m_O'} D^{(l_O)}_{m_O m_O'}(g) \mathcal{L}^{(l_O)}_{acm_O'}\big(\vec{r}_a, V^{(l_I)}_{acm_I}\big).
\end{align*}

\begin{figure}[ht]
\vskip 0.2in
\begin{center}
\centerline{\includegraphics[width=\columnwidth]{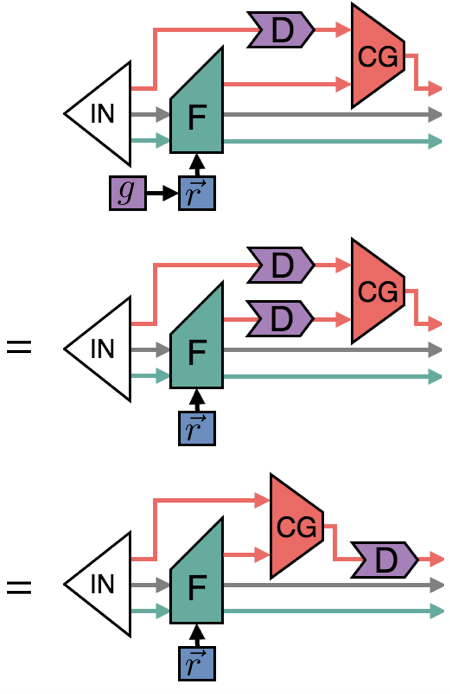}}
\caption{Diagrammatic proof of point convolution rotation equivariance.}
\label{fig:cg2}
\end{center}
\vskip -0.2in
\end{figure}

\section{Details for gravitational accelerations and moment of inertia tasks \label{sec:gravity_details}}

\subsection{Moment of inertia radial functions}

We can write the moment of inertia tensor as
\[I_{ij} := \sum_{a\in S} m_a T_{ij}(\vec{r}_a)\]
where
\[T_{ij}(\vec{r}) := R^{(0)}(r)\delta_{ij} + R^{(2)}(r)\left(\hat{r}_i\hat{r}_j - \frac{\delta_{ij}}{3}\right)\]
The expression that $R^{(2)}$ is multiplying is the 3D symmetric traceless tensor, which can be constructed from the $l=2$ spherical harmonic.  To get agreement with the moment of inertia tensor as defined in the main text, we must have
\[R^{(0)}(r) = \frac{2}{3}r^2 \qquad R^{(2)}(r) = -r^2\]

\begin{figure}[ht]
\vskip 0.2in
\begin{center}
\centerline{\includegraphics[width=\columnwidth]{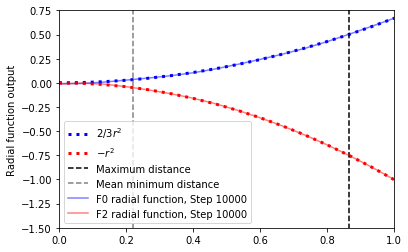}}
\caption{Radial function learned by $l=0$ and $l=2$ filters for moment of inertia toy dataset. The filters learn the analytical radial functions. For a collection of randomly generated point sets, the mean minimum distance is the average of the minimum distance between points in each point set. Distances smaller than the mean minimum distance might not have been seen by the network enough times to correct the radial filter.}
\label{fig:moment_radial}
\end{center}
\vskip -0.2in
\end{figure}

Figure~\ref{fig:moment_radial} shows the excellent agreement of our learned radial functions for filters $l=0$ and $l=2$ to the analytical solution.

\subsection{Point generation details and radial hyperparameters}
The number of points is uniformly randomly selected from 2 through 10. The masses are scalar values that are randomly chosen from a uniform distribution from 0.5 to 2.0. The coordinates of the points are randomly generated from a uniform distribution to be inside a cube with sides of length 4 for gravity and 1 for moment of inertia.

We use 30 Gaussian basis functions whose centers are evenly spaced between 0 and 2. We use a Gaussian variance that is one half the distance between the centers.  We use a batch size of 1. For the test set, we simply use more randomly generated points from the same distribution.

We note that $-1/r^2$ diverges as $r \to 0$. We choose a cutoff minimum distance at 0.5 distance because it is easy to generate sufficient examples at that distance with a few number of points per example. If we wanted to properly sample for closer distances, we would need to change how we generate the random points or use close to 1000 points per example.

\section{Proof of weighted point-averaging layer equivariance}

Let $S$ be the set of points (not including the missing point at $\vec{M}$) with locations $\vec{r}_a$.  Suppose that the output of the network is a scalar and a vector $\vec{\delta}_a$ at each point in $S$.  We take the softmax of the scalars over $S$ to get a probability $p_a$ at each point.  Define the votes as $\vec{v}_a := \vec{r}_a + \vec{\delta}_a$, so the guessed point is \[\vec{u} := \sum_{a\in S} p_a \vec{v}_a\]  This is the first operation that we have introduced that lacks manifest translation equivariance because it uses $\vec{r}_a$ by itself instead of only using $\vec{r}_{a} - \vec{r}_b$ combinations.  We can show that $\vec{r}_a \mapsto \vec{r}_a + \vec{t}$ implies 
\[\vec{u} \mapsto \sum_{a\in S} \left[p_a (\vec{r}_a + \vec{t}) + p_a \vec{\delta}_a\right] = \vec{u} + \vec{t}\] 
because the $p_a$ sum to~$1$.  This voting scheme is also rotation-equivariant because it is a sum of 3D vectors.  The loss function \[\text{loss} = (\vec{u} - \vec{M})^2\] is translation-invariant because it is a function of the difference of vectors in 3D space and rotation-invariant because it is a dot product of vectors.

\section{Missing point task accuracies and MAE by epoch}

\begin{figure}[ht]
\vskip 0.2in
\begin{center}
\centerline{\includegraphics[width=\columnwidth]{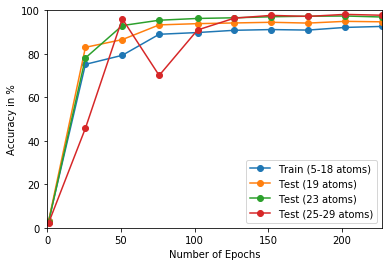}}
\caption{Accuracy of missing point task by epoch of training}
\label{fig:accuracy_by_epoch}
\end{center}
\vskip -0.2in
\end{figure}

\begin{figure}[ht]
\vskip 0.2in
\begin{center}
\centerline{\includegraphics[width=\columnwidth]{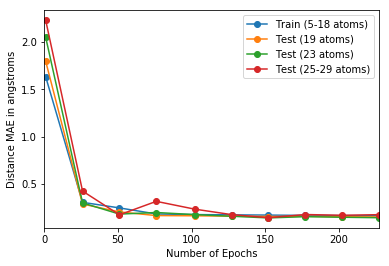}}
\caption{Distance MAE of missing point task by epoch of training}
\label{fig:mae_by_epoch}
\end{center}
\vskip -0.2in
\end{figure}

In Table~\ref{table:missing_point_by_atom}, we give the prediction accuracy and MAE on distance for the missing point task broken down by atom type. There are 1,000 molecules in each of the train and test sets; however, when comparing results by atom type, the relevant number to compare is the number of examples where a specific atom type is removed. In Figure~\ref{fig:accuracy_by_epoch} and Figure~\ref{fig:mae_by_epoch}, we give the accuracy and distance MAE for the missing point task as a function of the number of training epochs (Tables~\ref{table:missing_point}~and~\ref{table:missing_point_by_atom} contain the results after 225 epochs). 

\begin{table}[t]
\caption{Performance on missing point task by atom type}
\label{table:missing_point_by_atom}
\vskip 0.1in
\centering
\begin{small}
\begin{tabular}{ l S S S }
\toprule
Atoms & {\makecell{Number of\\ atoms with \\ given type\\ in set}} & {\makecell{Accuracy (\%) \\ ($\le 0.5$~\AA \\ and atom type)}} & {\makecell{Distance \\ MAE in \AA}} \\
\midrule
Hydrogen & & & \\
\midrule
5-18 (train) & 7207 & 94.6 & 0.16 \\
19 & 10088 & 93.2  & 0.16\\
23 & 14005 & 96.7 & 0.14\\
25-29 & 16362 & 97.7 & 0.15\\
\midrule
Carbon & & & \\
\midrule
5-18 (train) & 5663 & 94.3 & 0.16 \\
19 & 6751 & 99.9  & 0.10\\
23 & 7901 & 100.0 & 0.11\\
25-29 & 8251 & 99.7 & 0.17\\
\midrule
Nitrogen & & & \\
\midrule
5-18 (train) & 1407 & 74.8 & 0.16 \\
19 & 616 & 74.7 & 0.18\\
23 & 37 & 81.1 & 0.19\\
25-29 & 16 & 93.8 & 0.26\\
\midrule
Oxygen & & & \\
\midrule
5-18 (train) & 1536 & 83.3 & 0.17 \\
19 & 1539 & 80.2  & 0.21\\
23 & 1057 & 68.0 & 0.20\\
25-29 & 727 & 60.1 & 0.21\\
\midrule
Fluorine & & & \\
\midrule
5-18 (train) & 50 & 0.0 & 0.18 \\
19 & 6 & 0.0 & 0.07 \\
23 & 0 & & \\
25-29 & 0 & & \\
\bottomrule
\end{tabular}
\end{small}
\vskip -0.1in
\end{table}


\end{document}